\documentclass[
]{ceurart}

\usepackage{listings}
\usepackage{csquotes}
\usepackage{comment}
\renewcommand{\mkbegdispquote}[2]{\itshape}
\begin{document}

\copyrightyear{2022}
\copyrightclause{Copyright for this paper by its authors.
  Use permitted under Creative Commons License Attribution 4.0
  International (CC BY 4.0).}

\conference{In A. Martin, K. Hinkelmann, H.-G. Fill, A. Gerber, D. Lenat, R. Stolle, F. van Harmelen (Eds.), 
Proceedings of the AAAI 2022 Spring Symposium on Machine Learning and Knowledge Engineering for Hybrid Intelligence (AAAI-MAKE 2022), 
Stanford University, Palo Alto, California, USA, March 21–23, 2022.}

\title{Do it Like the Doctor: How We Can Design a Model That Uses Domain Knowledge to Diagnose Pneumothorax}
% Other Ideas
% Using Domain Knowledge to Design a Model to Diagnose Pneumothorax
% A Method on Eliciting Expert Domain Knowledge to Design a Model to Diagnose Pneumothorax


\author[1]{Glen Smith}[
email=gs675@drexel.edu,
]
\author[1]{Qiao Zhang}[
email=qz99@drexel.edu,
]
\author{Christopher J. MacLellan}[
email=cm3786@drexel.edu]
\address{Drexel University, Philadelphia, Pennsylvania, 19104, United States}
\address[1]{These authors contributed equally to this work.}
% \address[2]{Christopher J. MacLellan is the corresponding author.}

% \author[1,2]{Dmitry S. Kulyabov}[%
% orcid=0000-0002-0877-7063,
% email=kulyabov-ds@rudn.ru,
% url=https://yamadharma.github.io/,
% ]
% \address[1]{Peoples' Friendship University of Russia (RUDN University),
%   6 Miklukho-Maklaya St, Moscow, 117198, Russian Federation}
% \address[2]{Joint Institute for Nuclear Research,
%   6 Joliot-Curie, Dubna, Moscow region, 141980, Russian Federation}

% \author[3]{Ilaria Tiddi}[%
% orcid=0000-0001-7116-9338,
% email=i.tiddi@vu.nl,
% url=https://kmitd.github.io/ilaria/,
% ]
% \address[3]{Vrije Universiteit Amsterdam, De Boelelaan 1105, 1081 HV Amsterdam, The Netherlands}

% \author[4]{Manfred Jeusfeld}[%
% orcid=0000-0002-9421-8566,
% email=Manfred.Jeusfeld@acm.org,
% url=http://conceptbase.sourceforge.net/mjf/,
% ]
% \address[4]{University of Skövde, Högskolevägen 1, 541 28 Skövde, Sweden}

\begin{abstract}
  Computer-aided diagnosis for medical imaging is a well-studied field that aims to provide real-time decision support systems for physicians. These systems attempt to detect and diagnose a plethora of medical conditions across a variety of image diagnostic technologies including ultrasound, x-ray, MRI, and CT. When designing AI models for these systems, we are often limited by little training data, and for rare medical conditions, positive examples are difficult to obtain. These issues often cause models to perform poorly, so we needed a way to design an AI model in light of these limitations. Thus, our approach was to incorporate expert domain knowledge into the design of an AI model. We conducted two qualitative think-aloud  studies with doctors trained in the interpretation of lung ultrasound diagnosis to extract relevant domain knowledge for the condition Pneumothorax. We extracted knowledge of key features and procedures used to make a diagnosis. With this knowledge, we employed knowledge engineering concepts to make recommendations for an AI model design to automatically diagnose Pneumothorax.
\end{abstract}


\maketitle

\section{Introduction}

When building artificial intelligence (AI) models with limited data, we are often concerned with issues of low performance. This may be due to not having enough data for the model to effectively learn the relationships between the input and the output. Another example may be over-fitting, where the model learns the noise and nuance of the training data well, but performs poorly on unseen data.  Further, for many classification tasks, we often lack sufficiently balanced datasets, which additionally leads to lowered model performance. 

One such way to mitigate some of these issues is by incorporating subject matter expert knowledge, called ``domain knowledge'' \cite{alexander1992domain}, into the design of an AI model. In essence, we can ask an expert ``how would you accomplish this task?'' and extract key steps, milestones, and outcomes that we should consider in the model design. This approach allows us to build more targeted and robust models that focus on specific, expert-defined features. In this research, we present two studies whose aims are to qualitatively extract domain knowledge from physicians in the POCUS (point-of-care ultrasound) domain for the task of detecting and diagnosing Pneumothorax.

An AI system capable of diagnosing Pneumothorax has many applications, one of which is assisting medics in the military. Battlefield medics are often required to make real-time diagnoses of multiple conditions in stressful, high-stakes environments. Ultrasound is one of the recommended ways for these medics to diagnose Pneumothorax, but training to interpret ultrasounds can be resource heavy. Thus, there is a need for automated support systems that can assist in making these diagnoses.

Previous studies show that incorporating expert knowledge into an AI model's design can produce higher and more robust performance. \cite{donahue2011annotator, sharma2018learning, sharma2016towards}.
% [Donahue, Grauman Annotator Rationales for Visual Recognition]
% [Sharma, Bilgic 2017 Learning with Rationales for Document Classification]
% [Sharma, Bilgic 2016 Towards Learning with Feature-Based Explanations for Document Classification]
In this work, we analyze the process subject matter experts use to diagnose Pneumothorax in lung ultrasound videos and determine which features and artifacts in the videos are considered, the order in which they are considered, and the relative importance of these features. 

We considered two main model design inspirations for our studies, robustness and user confidence. With a limited training dataset (62 ultrasound videos in our case), it is difficult for a machine learning model to extract meaningful distinctions between features to make an accurate diagnosis. To address this issue, we aimed to extract two kinds of domain knowledge from our medical experts: knowledge of key features and inference knowledge. Having a method to elicit important features from experts lets us develop a model design that hones in on those attributes, which creates a more robust model \cite{frank2021integrating}.
% [Frank et. al. Integrating Domain Knowledge into Deep Networks for Lung Ultrasound with Applications to COVID-19]
Further, by extracting the procedures for diagnosing Pneumothorax, we can similarly develop inference rules to incorporate into a model. This domain knowledge should result in AI models that require fewer training instances, less training time, and provide higher performance overall.

In this paper, we first provide some background on \emph{cognitive task analysis} and the \emph{think-aloud} method, which is the type of cognitive task analysis we employ for our work. We then present a brief overview of the POCUS domain to establish the context and need for AI technology to diagnose Pneumothorax. Next, we present two studies, that outline how we elicited domain knowledge from two medical experts and the results of these sessions. In the discussion, we provide recommendations for the design of an AI diagnosis system based on our findings from these studies. Finally, we discuss related works, our unique contributions to the space of knowledge engineering for designing AI models to help diagnose Pneumothorax, and conclude with ongoing and future work.

\section{Background}

\subsection{Cognitive task analysis}

Cognitive Task Analysis (CTA) is the process of extracting knowledge, processes, and patterns from individuals as they attempt to evaluate a scenario in a particular domain or problem space \cite{zsambok2014naturalistic}.
% [NDM, Zsambok, Klein Naturalistic Decision Making]
With this analysis, researchers can evaluate the cognitive load involved in performing a task, determine the knowledge and procedures an individual uses to complete a task, and the knowledge required to determine when a task is complete. Most often, subject matter experts are the focus of such studies as they usually contain the breadth of knowledge required to navigate the problem space or their respective domains. However, in some cases it may be desirable to study how a novice might tackle a similar problem and what errors are present in their processes. The use of the CTA method is therefore to develop systems and processes based on expert and/or novice knowledge. Some examples of these systems and processes include training programs\cite{seamster2017applied}, and
% [Redding and Seamster, 2017 Applied Cognitive Task Analysis in Aviation]
medical procedures \cite{sullivan2007use}.
% [cite, mb Sullivan et. al. The use of cognitive task analysis to improve the learning of percutaneous tracheostomy placement]

There are many CTA methods used to facilitate knowledge engineering and each varies in terms of their knowledge representation, how information is extracted from participants, and the types of tasks the methods work best for. These facets are further divided into their own categories of methods for knowledge elicitation and representation \cite{klein20014}.
% [Klein, Militello 2001 SOME GUIDELINES FOR CONDUCTING A COGNITIVE TASK ANALYSIS].
When used for designing training systems, CTA is used two-fold, to first define the difficulty of the task in the problem space, determine common errors encountered by expert in the field, and produce ways to mitigate those errors \cite{woods1989cognitive}.
% [Roth, Woods, 1989 COGNITIVE   TASK   ANALYSIS:   AN   APPROACH   TO   KNOWLEDGE   ACQUISITION   FOR   INTELLIGENT  SYSTEM    DESIGN ]
This problem formulation stage focuses on the theory of the task and the cognitive processes required by participants to address it. The next stage examines the application of stage one, to determine technical requirements, reasons for performance issues, how a system might solve them, and what computational tools to employ \cite{woods1989cognitive}.
% [Roth, Woods, 1989 COGNITIVE   TASK   ANALYSIS:   AN   APPROACH   TO   KNOWLEDGE   ACQUISITION   FOR   INTELLIGENT  SYSTEM    DESIGN ]. 

Some specific CTA methodologies employed for knowledge elicitation include unstructured interviews, critical decision analyses, direct observation and questioning, and simulations \cite{klein20014}.
% [Klein, Militello 2001, Some guidelines for conducting a CTA].
In these approaches, researchers will either ask participants a series of questions, allowing for feedback and revising of further questions, or observe participants as they attempt to solve a problem or walk through a process, usually with some verbal component from either party. Often, a combination of some of these methods produce a robust qualitative method for extracting domain knowledge from subject matter experts. 

According to \cite{zsambok2014naturalistic}
% [Zsambok and Klein, 1997 Naturalistic Decision Making] 
the CTA method is most effectively employed when the analysis contains ``complex, ill-structured tasks'' that may have multiple solutions, ``dynamic, uncertain, and real-time environments'' that the individual must navigate, or some level of multi-tasking where decisions are based on a variety of simultaneous conditions. Furthermore, they go on to cite that CTA is appropriate when the problem consists of complex perceptual learning and pattern recognition. All of these conditions apply to our use case, where battlefield medics must make real-time diagnoses of Pneumothorax in often stressful environments. Battlefield medics are often multi-tasking to tend to various injuries, and the constantly changing environment lends itself to uncertainty and places a load on the medics to make real-time decisions. This makes CTA a suitable method to extract physician knowledge for the purpose of developing an intelligent AI system to diagnose Pneumothorax. As explored in our two studies conducted with our physicians, we employed the Think-Aloud CTA strategy to inform the design of an AI model to diagnose Pneumothorax.

\subsection{Think-Aloud Analysis}

Think Alouds (TAs) are one specific type of cognitive task analysis that asks participants to verbalize each step in their problem-solving process \cite{fonteyn1993description}.
% [Fonteyn et. al, Desc. of Think Aloud method & Protocol Anal.]
In general, researchers conduct one-on-one sessions where they give a participant a problem to solve. The participant outlines their solution to the problem either ``retrospectively'' – thinking back on a previously solved problem – or ``concurrently'', where the problem is solved real-time with an accompanied verbalization \cite{ericsson1980verbal}.
% [Ericsson & Simon, 1980, Verbal reports as data]
Since the verbalization, or the ``verbal report'', is the output data of the study, the study is transcribed and often recorded for later review. Transcripts are then analyzed to extract entities, concepts, decision flows, and more. 

We make use of the Think-Aloud method for our studies for a myriad of reasons. Compared to other qualitative methods, the Think-Aloud method has been shown to be robust against various ``errors'' that would otherwise invalidate the data obtained from the process \cite{van1994think}.
% [van Someren et. al, The Think Aloud method: a practical appr. To modelling cognitive processes]
More precisely, Someren et. al. state that the act of verbalization can help resolve memory errors and allow for a slower, yet more complete cognitive flow. Think-Aloud studies are best utilized in scenarios of small participant sample sizes and simulation environments, where a participant works through a controlled ``real-world'' example of the problem \cite{fonteyn1993description}.
% [Fonteyn et. al, Desc. of Think Aloud method & Protocol Anal.]

\section{POCUS Task: Pneumothorax Diagnosis}

\begin{figure}[!htbp]
\centering
\includegraphics[width=6cm]{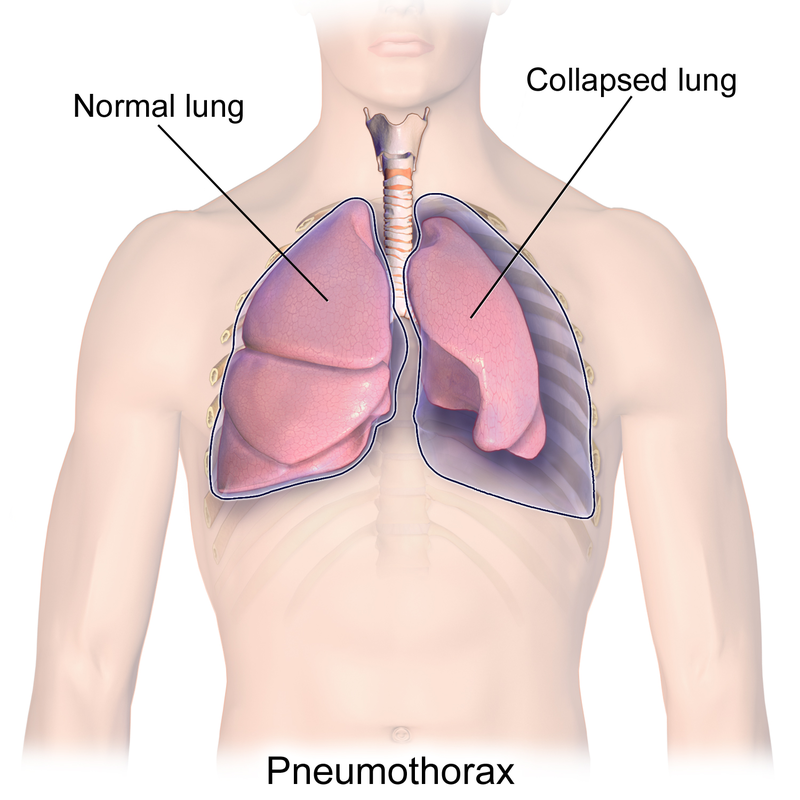}
\caption{An anatomical diagram demonstrating the difference between a normal lung and a collapsed lung. \cite{blausen2014medical}}
\label{fig:PTX}
\end{figure}

% introduce POCUS and POCUS AI, we are building an AI model to do video classification
POCUS, which stands for Point-Of-Care Ultrasound, involves the use of an ultrasound device to answer specific diagnostic questions and to assess real-time physiological responses to treatment \cite{damodaran2020artificial}. In the scope of POCUS AI, we aim to use this research to build a system that accepts ultrasound videos as input and classifies whether the video is an example of Pneumothorax \cite{DARPAnews}. A Pneumothorax, also known as collapsed lung or dropped lung, is the entry of air into the pleural space (the space between the lungs and chest wall). When air enters this area, the lung loses contact with the inside of the chest and ``drops'' down \cite{sahn2000spontaneous}. Figure \ref{fig:PTX} is an anatomical diagram that compares a normal lung with one conditioned with Pneumothorax. In most cases, a Pneumothorax is caused by a traumatic injury, such as a rib fracture or penetrating injury (stab or gunshot wound) that causes damage to the lung or chest \cite{choi2014pneumothorax}. 

% use advanced artificial intelligence techniques to diagnose Pneumothorax by producing code and develop a system demonstrating the feasibility interpretation of POCUS across multiple applications.

\begin{figure}[!htbp]
\includegraphics[width=7cm]{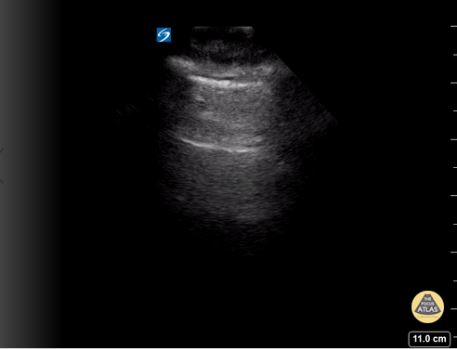}
\includegraphics[width=7cm]{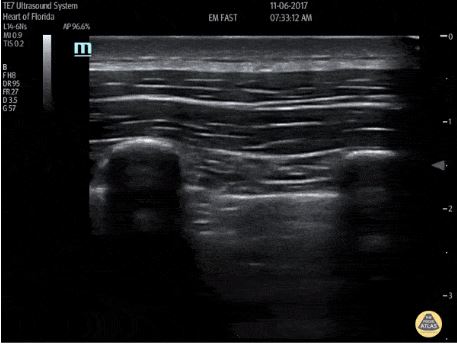}
\caption{Two Ultrasound snapshots labeled with ``sliding'' (left) and ``no sliding'' (right). [Public domain], via the POCUS ATLAS. (\url{https://www.thepocusatlas.com/lung/5l9jgyaszu0othj5tidg0miqxkmvyv}) provided by Hannah Kopinski (MS4), Dr. Lindsay Davis and Matthew Riscinti, (\url{https://www.thepocusatlas.com/lung/no-lung-sliding}) provided by Francisco Norman.}
\label{fig:two_us}
\end{figure}

% the data we are using
The video data used for this study are provided by Brooke Army Medical Center (BAMC). Additionally, we utilized open-source ultrasound video data from the POCUS ATLAS \cite{PocusAtlas} (examples are shown in Figure \ref{fig:two_us}). For the BAMC data, we have 32 videos labeled with ``sliding'' and 30 videos labeled with ``no sliding'', where a label of ``no sliding'' is indicative of pneumothorax. Each video is a 3 second short clip that contains 20 frames per second. While some of the ultrasound videos were from the same patient, each of the videos has a unique file name. Examples of ``sliding'' and ``no sliding'' snapshots from the POCUS ATLAS videos are provided in Figure \ref{fig:two_us}.

% capturing domain experts' knowledge
Though we have limited training data for a complex machine learning system, the video data provide adequate information for experienced medical experts to diagnose Pneumothorax. Therefore, we were able to employ Think-Aloud analyses with our medical experts to aid in the design of a decision support system \cite{xie2021survey}. In this work, we captured the cognitive reasoning processes that the medical experts, identified in our studies with the pseudonyms Alex and Victor, use when diagnosing Pneumothorax. In the next two sections, we present the motivations, designs, analyses, and results of our two studies.

\section{Study 1: Knowledge for Making Diagnoses}

\subsection{Motivation}
To diagnose a condition such as Pneumothorax using ultrasounds, doctors must be aware of the condition's specific characteristics. They are knowledgeable of the features and artifacts that confirm, reject, or even make uncertain whether a patient has this condition. Further, it is expected that they can explain what these features are and why these features are relevant to a diagnosis. If we can elicit this knowledge as researchers, we can use it to design systems that can detect these specific features, make diagnoses using a similar process to the doctors, and generate explanations in terms that doctors can understand.

Therefore, the motivation of this study is to examine the first of the aforementioned expectations: defining the features of Pneumothorax in ultrasound videos. When a doctor makes a diagnosis, there is much we can learn about their process that would be useful for designing and building a robust AI model to detect features in an ultrasound video. We designed a Think-Aloud study to extract this information from two doctors trained in the use and interpretation of point-of-care ultrasound captures. As we will see in the design and analysis sections, we were able to determine both the features of the ultrasound as well as the medical concepts they employed to make the diagnosis.

\subsection{Design}
For each doctor, we conducted two Think-Aloud sessions where we prompted them to ``make a diagnosis of Pneumothorax''. In the first session with each doctor, we asked them to diagnosis Pneumothorax within six ultrasound videos. This session served as a ``warmup'' study to familiarize the doctors with the Think-Aloud format and from which we might extract preliminary features of their diagnoses. The second session with each doctor repeated the process with six additional lung ultrasound videos. This approach of pairing a study with a ``warmup'' study is suggested by Someren et. al. \cite{van1994think}. The ultrasound videos used for the warmup sessions were open-source data from the POCUS ATLAS \cite{PocusAtlas}, and the videos used for the second set of sessions were from a dataset from the Brooke Army Medical Center.

Of the six videos in each session, three were examples of Pneumothorax and three were examples of non-Pneumothorax. We asked the doctors to verbalize their thinking process as they worked through the diagnosis. The videos were presented in a random order and each doctor was unaware of the true diagnosis of all videos until the end of the study. The Think-Aloud sessions occurred over Zoom and remote control of the computer's mouse was also provided so that the doctors could control the videos and visually annotate them. 

With the doctor's consent, we recorded the entirety of each session for further review. During the Think-Aloud sessions, we were careful not to say things to the doctors to influence their decision making. Our interactions only consisted of a prompt to continue speaking if there was a pause of more than 3-5 seconds. This intervention was instrumental in ensuring all steps of the diagnosis process were verbalized and recorded. At the end of the session, we reviewed some keywords and concepts by asking the doctors to clarify some of their results. This ensured we captured the full breadth of knowledge presented in the Think Aloud. Finally, session transcripts were automatically generated from the recordings and these, along with the recordings, were used during our analysis.

It should be noted that in the original BAMC dataset, labels of ``Sliding'' and ``No Sliding'' were used as proxies for ``No Pneumothorax'' and ``Pneumothorax'' labels, respectively. In this context, sliding and no sliding refers to the movement of the pleural line, a key anatomical feature that doctors look for when diagnosing Pneumothroax in lung ultrasounds. Whether sliding is visible in the pleural line is one of the strongest indicators of their being no Pneumothorax. However, since sliding is not an entirely definitive proxy for a Pneumothorax diagnosis, this presented one limitation of the data. We conducted our analysis with this limitation in mind. Our study, including the use of the existing ultrasound data, was reviewed and approved by Drexel's Institutional Review Board.

\subsection{Analysis}

Alex and Victor made correct diagnosis on each of the six videos in Session 1 and Session 2. After we conducted our Think-Aloud studies with each doctor, we analyzed the transcripts to extract keywords and concepts that the doctors used. Transcripts were automatically generated by zoom, and prior to our analysis we cleaned the scripts by correcting grammar and punctuation, clarifying any misspelled vocabulary, and putting the dialogue in a standard format that grouped content by speaker and line. We then determined key themes and features present in the transcripts. Here is a snippet of one of the doctor's dialogue: 
\begin{displayquote}
The issue here is that everything is shifting, and if I look at the line here, here, here, and here, I don't see independent movement. I don't see vertical artifacts.
\end{displayquote}
From this quote, we extracted two of our concepts seen in Tables \ref{tbl:study_1_qiao} and \ref{tbl:study_1_glen}, \emph{movement} and \emph{vertical artifacts}.

%- could display image of videos from public domain 
\subsection{Results}
We observed twelve key concepts in sessions one (see Table \ref{tbl:study_1_qiao}) and two (see Table \ref{tbl:study_1_glen}). For each video, we label which doctors mention which concepts, using the square for Alex and triangle for Victor. For these studies, we simply mark if a concept is present in their discussion, but do not notate the number of times the concept is mentioned.

Through our analysis, we determined that there is a further distinction between the concepts. We define ``features'' of the ultrasound, which are objects or regions in the video, and ``visual characteristics'', which are characteristics the features exhibit, such as a type of movement. We noticed that some features and characteristics are discussed with higher frequency across all the videos compared to others such as \emph{pleural line} and \emph{movement}.

%lines from Qiao
We summarized the keywords and concepts both doctors mentioned during the two sessions in Table \ref{tbl:study_1_summary}. The \emph{pleural line} is generally agreed upon by the two doctors as a critical feature to conduct analysis in both ``sliding'' and ``no sliding'' scenarios. Alex always used anatomical landmarks such as rib or muscle to locate the pleural line while Victor did not; we believe that this could be explained as personal diagnosing preference since anatomical landmarks are not a cause or sign of Pneumothorax. Lastly, Alex mentioned ``lung pulse'' and ''vertical reverberation'' when making diagnosis on ''no sliding'' videos, since these two features only appear in ultrasounds where the patients have a high likelihood of Pneumothorax.

As we will see in the discussion section, we are able to extract a large amount of value from this study and we present some ways this information can be utilized to design an AI system to diagnose Pneumothorax. 

% add a row to describe the label of each video
% Square for Alberto and triangle for Vincent
\begin{table}[!htbp]
    \caption{Keywords summarized from Study 1 Session 1 ($\square$ represents Alex and $\triangle$ represents Victor)}
    \centering
    \begin{tabular}{|p{3.5cm}||p{1.5cm}|p{1.5cm}|p{1.5cm}|p{1.5cm}|p{1.5cm}|p{1.5cm}|}
    \hline
        & Video 1 & Video 2 & Video 3 & Video 4 & Video 5 & Video 6\\
        {Video labels} & Sliding & No Sliding & No Sliding & Sliding & No Sliding & Sliding\\
        \hline
        \hline
        {\bf Ultrasound Features} & & & & & &\\
         \hline
        Pleural line & $\square$ & $\square$ $\triangle$ & $\square$ $\triangle$ & $\square$ $\triangle$ & $\square$ $\triangle$ & $\square$ $\triangle$\\
        \hline
        Anatomical landmarks &$\square$ $\triangle$ & $\square$ & $\square$ & $\square$ $\triangle$ & $\square$ $\triangle$ & $\square$\\
        \hline
        B line &$\triangle$ & & $\triangle$ & & &\\
        \hline 
        Z line & $\square$ $\triangle$ & & & & &\\
        \hline
        A line & & $\square$ $\triangle$ & & & &\\
        \hline
        \hline
        {\bf Visual Characteristics} & & & & & &\\
        \hline
        Lung pulse & & $\square$ & $\square$ & $\square$ & $\square$ &\\
        \hline
        Lung point(s) & & $\square$ $\triangle$ & $\square$ $\triangle$ & & $\square$ &\\
        \hline
        Movement & $\square$ & $\square$ $\triangle$ & $\triangle$ & $\triangle$ & $\square$ $\triangle$ & $\square$ $\triangle$\\
        \hline
        Vertical artifacts & $\square$ $\triangle$ & $\square$ & & & &\\
        \hline
    \end{tabular}
    
    \label{tbl:study_1_qiao}
\end{table}

% Check scripts to confirm what the term ``movement'' refers to

\begin{table}[!htbp]
    \caption{Keywords summarized from Study 1 Session 2 ($\square$ represents Alex and $\triangle$ represents Victor)}
    \centering
    \begin{tabular}{|p{3.5cm}||p{1.5cm}|p{1.5cm}|p{1.5cm}|p{1.5cm}|p{1.5cm}|p{1.5cm}|}
    \hline
        & Video 7 & Video 8 & Video 9 & Video 10 & Video 11 & Video 12\\
        {Video labels} & Sliding & Sliding & No Sliding & Sliding & No Sliding & No Sliding\\
        \hline
        \hline
        {\bf Ultrasound Features} & & & & & &\\
        \hline
        {Pleural line} & $\square$ $\triangle$ & $\square$ $\triangle$ & $\square$ $\triangle$ & $\square$ $\triangle$ & $\square$ $\triangle$ & $\square$ $\triangle$\\
        \hline
        {B line} & $\square$ & $\triangle$ & $\triangle$ & & & $\square$\\
        \hline
        {Z line} & & $\triangle$ & & & &\\
        \hline
        \hline
        {\bf Visual Characteristics} & & & & & &\\
        \hline 
        {Lung pulse} & & & $\triangle$ & & & $\square$ $\triangle$\\
        \hline
        Movement & $\square$ & $\square$ & $\square$ $\triangle$ & $\square$ $\triangle$ & $\triangle$ & $\square$\\
        \hline
        Acoustic shadowing & $\square$ & & $\triangle$ & $\square$ & $\square$ &\\
         \hline
        Horizontal sliding & & & $\square$ $\triangle$ & $\triangle$ & & $\square$\\
        \hline
        Vertical reverberation & $\square$ $\triangle$ & $\triangle$ & $\square$ & $\square$ & $\square$ & $\square$\\
        \hline
    \end{tabular}
    \label{tbl:study_1_glen}
\end{table}

\begin{table}[!htb]
    \caption{Common features summarized from Session 1 and Session 2 in Study 1 ($\square$ represents Alex and $\triangle$ represents Victor)}
    \centering
    \begin{tabular}{|p{3.5cm}||p{1.5cm}|p{1.5cm}|}
    % \begin{tabular}{|c||c|c|c|c|}
    \hline
        {Video labels} & Sliding & No Sliding\\
        % & Pleural line & Anatomical landmarks & Lung pulse & Vertical reverberation\\
    \hline
    \hline
    {\bf Ultrasound Features} & &\\
    \hline
    Pleural line & $\square$+6, $\triangle$+5 & $\square$+6, $\triangle$+5\\
    \hline
    B line & $\square$+1, $\triangle$+2 & $\square$+1, $\triangle$+2\\
    \hline
    Z line & $\square$+1, $\triangle$+2 &\\
    \hline
    \hline
    {\bf Visual Characteristics} & &\\
    \hline
    Lung pulse & $\square$+1 & $\square$+4, $\triangle$+2\\
    \hline
    Movement & $\square$+5, $\triangle$+3 & $\square$+5, $\triangle$+4\\
    \hline
    Vertical reverberation & $\square$+3, $\triangle$+3 & $\square$+4\\
    \hline
    \end{tabular}
    \label{tbl:study_1_summary}
\end{table}

% \begin{table}[!htb]
%     \caption{Common features summarized from Session 1 and Session 2 in Study 1 ($\square$ represents Alex and $\triangle$ represents Victor)}
%     \centering
%     \begin{tabular}{|p{2cm}||p{2.25cm}|p{3.5cm}|p{2.25cm}|p{3.5cm}|}
%     % \begin{tabular}{|c||c|c|c|c|}
%     \hline
%         & Pleural line & Anatomical landmarks & Lung pulse & Vertical reverberation\\
%     \hline
%     Sliding & $\square$ $\triangle$ & $\square$ & &\\
%     \hline
%     No sliding & $\square$ $\triangle$ & $\square$ & $\square$ & $\square$\\
%     \hline
%     \end{tabular}
%     \label{tbl:study_1_summary}
% \end{table}

\section{Study 2: Knowledge for Explaining Diagnoses}

\subsection{Motivation}
While we have collected some important ultrasound features and visual characteristics from Study 1 that can help with diagnosing Pneumothorax, we wanted to further explore how domain experts produce a reasonable explanation of already-diagnosed conditions in ultrasound videos. This would help us to design for better better explainability and transparency in an AI system. We conducted this second study with the main purpose of capturing the domain experts' cognitive reasoning process when they see an ultrasound video paired with a previous diagnosis and are asked to produce a reasonable explanation of the diagnosis. The main difference between study 1 and study 2 is that study 1 observed how the experts generated a diagnosis of Pneumothorax, while study 2 observes how the experts explain a diagnosis of Pneumothorax. In study 2, we wanted to understand if and how the reasoning process changes when asked to confirm or reject a predetermined diagnosis and whether there is variation between the experts' diagnosis processes.

% Upon completing Study 1, where we conducted two think-aloud sessions with domain experts and had them doing the diagnosis, we wanted to explore if the domain knowledge and the cognitive reasoning process they employed to do the diagnosis is different from what they used to explain the diagnosis made by other experts.

% - talk about Dr. Weber's hypothesis, additional knowledge should be needed
% find references for the hypothesis

\subsection{Design}
We selected four videos from the BAMC data, two of them were labeled with ``Sliding'' and the other two were labeled with ``No Sliding''. The four videos were collected from four different patients. Similar to Study 1, we showed the videos to the medical experts in random order, recorded the entire session, and transcribed the recordings. In contrast to study 1, we presented the four videos with the accompanying diagnosis labels to the medical experts. To better observe how they extract key features and generate reasonable explanations, we flipped the labels (e.g., a ``Sliding'' label would be shown as ``No Sliding'' and vice versa) of one ``Sliding'' video and one ``No sliding''video, thus providing two videos with the correct labels and the other two with incorrect labels. Ultimately, we have one video correctly labeled with ``Sliding'', one video correctly labeled with ``No Slding'', one video incorrectly labeled with ``Sliding'', and one video incorrectly labeled with ``No Sliding''.

The original ultrasound video labels and the labels that we showed the medical experts are displayed along with our analysis results in Table \ref{tbl:study_2}. Our study, including the use of the existing ultrasound data, was reviewed and approved by Drexel's institutional review board.

\begin{table}[!htb]
    \caption{Keywords summarized from Study 2 ($\square$ represents Alex and $\triangle$ represents Victor)}
    \centering
    % \begin{tabular}{|p{0.45\linewidth}||c|c|c|c|}
    \begin{tabular}{|p{5cm}||p{1.5cm}|p{1.5cm}|p{1.5cm}|p{1.5cm}|}
    \hline
        & Video 13 & Video 14 & Video 15 & Video 16\\
        \hline
        Original label & Sliding & No Sliding & No Sliding & Sliding\\
        Presented label & Sliding & Sliding & No Sliding & No Sliding\\
        \hline
        \hline
        {\bf Ultrasound Features} & & & &\\
        \hline
        Pleural line & $\square$ & $\square$ $\triangle$ & $\square$ & $\square$ $\triangle$\\
        \hline
        Anatomical landmark & $\square$ & $\square$ & $\square$ & $\square$\\
        \hline
        \hline
        {\bf Visual Characteristics} & & & &\\
        \hline
        Lung pulse & $\square$ $\triangle$ & & &\\
        \hline
        Lung point(s) & & $\square$ $\triangle$ & & $\square$ $\triangle$\\
        \hline
        Moving/Movement & $\square$ & $\square$ $\triangle$ & $\square$ $\triangle$ & $\square$ $\triangle$\\
        \hline
        Sliding & $\square$ & $\square$ $\triangle$ & & $\square$ $\triangle$\\
        \hline
        No sliding & $\square$ $\triangle$ & $\square$ $\triangle$ & $\square$ $\triangle$ & $\square$ $\triangle$\\
        \hline
        Vertical reverberation artifact & & $\square$ & & $\square$ $\triangle$\\
        \hline
    \end{tabular}
    \label{tbl:study_2}
\end{table}

\subsection{Analysis}
By providing two videos with the correct labels and the other two videos with the incorrect labels, we aimed to not only extract features that help explain the true diagnosis, but also cause the medical experts to question their diagnosis process and be more critical towards the demonstrated diagnosis results.

An interesting finding we have from this session is that both medical experts agreed to disagree with two videos. For Video 13, we showed them a video correctly labeled with ``sliding'', while both doctors mentioned that they would prefer recognizing the phenomenon as a lung pulse (a vertical motion of the pleura in sync with the cardiac rhythm \cite{volpicelli2012international}) instead of sliding. They agreed that this is suspicious of Pneumothorax, but were unable to definitively confirm it. For Video 14, we presented a video incorrectly labeled with ``sliding'' while the ground truth BAMC label was ``No Sliding''. Victor made an argument of a possibly incorrect diagnosis. Although he observed majority no-sliding, he stated that he would need more information to make the decision. Similar to Victor, Alex described the video to have 80\% no-sliding and 20\% sliding, and suspected Pneumothorax (which would be linked to no-sliding). Video 15 is correctly labeled with ``No Sliding'' and both experts made quick decisions to agree with the labeling. Video 16 is presented with the flipped label ``No Sliding'' and the ground truth BAMC label is ``Sliding''. Victor described what he saw as half sliding and half no sliding. He thought the video was suspicious of Pneumothorax (i.e., no sliding) but would need more information to make the diagnosis. Alex described what he observed as ``clearly sliding''.

Here is a snippet from one of the doctor's TAs:
\begin{displayquote}
So the first thing I want to identify...I want to identify the pleural line, and so to identify the pleural line, I identify a rib... And so I know that the line, that is just underneath. And then I look at the movement because the question is ``sliding'' or ``no sliding''. I want to see an independent movement of sliding that could be seen here. So this is not a Pneumothorax for sure.
\end{displayquote}
From this quote, we extracted three of the concepts displayed in Table \ref{tbl:study_2}, \emph{pleural line}, \emph{anatomical landmark} (referring to rib here), and \emph{movement}.

% - 2nd think aloud session (Qiao) => flipping the script: giving the doctors the opposite label (a false label) to extract what features help to explain the true diagnosis (makes them question their process and be more critical of it)
% - maybe discuss how flipping the labels affects the ability for each doctor to make diagnosis/identify features, ie. for each doctor, is there a difference in how a diagnosis is made when they're presented with the right label vs wrong label, and the same thing across both doctors
%confirmation bias
%data analysis

\subsection{Results}

Upon completing the second Think-Aloud study, we analyzed the doctors' diagnoses in terms of sliding/no-sliding, Pneumothorax/no-Pneumothorax, and suspicions of Pneumothorax/no-Pneumothorax. Table \ref{tbl:study_2_diagnosis} displays the original BAMC label, presented label, and the two medical experts' explanations for each video.

Compared to Study 1, where both doctors did not prioritize among the ultrasound features and visual characteristics, Alex used \emph{inference rules} to construct his explanations of the previously labeled lung ultrasound videos. His cognitive reasoning process could be divided into four steps:
\begin{enumerate}
    \item Alex stated that he always looks for pleural line first. To recognize the pleural line, he identifies anatomical landmarks such as ribs and muscle.
    \item Next, he examined if there is any independent movement in the pleural line. If sliding is present along the entire pleural line, then he can confidently make a diagnosis of no Pneumothorax.
    \item If he did not see sliding or only saw partial sliding, then he would look for lung pulse. Recognizing pulse would lead to a conclusion of no Pneumothorax.
    \item Lastly, Alex would look for vertical artifacts. If there are vertical artifacts, then there is no Pneumothorax. However, if no vertical artifacts are observed, then more information is needed to make a definitive decision. 
\end{enumerate}
There's still a possibility that he cannot make a diagnosis after considering these four features sequentially. Poor image resolution or only seeing part of the lung were the common issues preventing him from making a diagnosis. To make a clinical decision in these cases, more ultrasound data would need to be collected to support decision making.

% - present the individual differences in a paragraph, show in a table \\
% - table would be: each row is a different concept that we extracted, each column is for each SME and we show if the doctors agreed on the same concept \\
% - is there anything different in study 2 from study 1 \\
% - Moving and sliding are used describing different things
% - Dr. Chan thinks that identifying ``sliding'' and ``no sliding'' is the first step of diagnosing Pneumothorax
% - The diagnosis is not a final decision, but hypothesis that doctors are making.
% for the videos with flipped labels, two doctors have different opinions

\begin{table}[!htb]
    \caption{Diagnosis analysis for Study 2 ($\square$ represents Alex and $\triangle$ represents Victor)}
    \centering 
    % \begin{tabular}{|p{0.45\linewidth}||c|c|c|c|}
    \begin{tabular}{|p{5cm}||p{1.5cm}|p{1.5cm}|p{1.5cm}|p{1.5cm}|}
    \hline
        & Video 13 & Video 14 & Video 15 & Video 16\\
        \hline
        Original label & Sliding & No Sliding & No Sliding & Sliding\\
        Presented label & Sliding & Sliding & No Sliding & No Sliding\\
        \hline
        \hline
        {\bf Explanation} & & & &\\
        \hline
        Sliding & & & & $\square$\\
        \hline
        Reduced sliding & $\square$ & & &\\
        \hline
        No sliding & $\triangle$ & & $\square$ $\triangle$ &\\
        \hline
        Pneumothorax & & $\square$ & & $\square$ $\triangle$\\
        \hline
        No Pneumothorax & $\square$ & & &\\
        \hline
        Not sure about sliding/no sliding& & $\square$ $\triangle$ & & $\triangle$\\
        \hline
    \end{tabular}
    \label{tbl:study_2_diagnosis}
\end{table}

\section{Discussion}

From our Think-Aloud studies, we determined that there are various ways in which we can incorporate lung ultrasound domain knowledge into the design of an AI system for this task.

We know that in order for a system to be capable of diagnosing Pneumothorax, it must be able to detect features that are \emph{relevant} to the medical condition. One way researchers could achieve this is by building an object detection system that locates various features within the frames of an ultrasound video. However, not all features are relevant to Pneumothorax, so in order to construct a more robust and targeted model, researchers must be judicious in how certain features are weighted \cite{hwang2018usefulness}. From our studies, we can not only derive the features of most importance to the doctors, but also an approximation of relative weight based on the frequencies the features were discussed across the video samples. Thus, one such model design could be an object (or feature) detector parameterized by the relative ``weights'' of those features. For example, we might build an object detector to identify the pleural line in an ultrasound video, so that subsequent analysis can focus on this feature.

An object detector is an excellent start, but only considers features of the ultrasound, not characteristics of those features. From the results of studies one and two, we know that ``movement'' plays an important role in determining Pneumothorax. In fact, it was stated by one doctor that without movement, it would be impossible to make a diagnosis for Pneumothorax. Thus, the interpretation for a researcher would be that the model must not consider just one frame, but a series of sequential frames to determine the type of movement a feature exhibits. We, therefore, know that the type of model must consider multiple frames as an input to detect any movement. 

%Qiao
Study two presents further insights that can help construct an AI model to diagnose Pneumothorax. In addition to ``weighting'' the features, we introduce the idea of ``inference rules'' by prioritizing the features we extracted. Inference rules suggest an order of detecting/recognizing ultrasound features and relevant visual characteristics, and provide the AI model with more knowledge for making classifications between Pneumothorax and no Pneumothorax. Similar to traversing a decision tree, the AI system would have knowledge about which feature to detect first, and whether to make a classification at that point or move on to the next feature. This approach could contribute to the accuracy of video classification, expedite the decision making process, and increase the transparency of AI-assisted image classification, which are crucial medical needs for battlefield diagnosis.

\section{Related Work}

% talked about our proposed approach of weighting features and using inference rules
There have been many other exciting works in the medical imaging space that seek to diagnose diseases using AI technology. In the scope of incorporating domain knowledge from medical experts, the closest work to ours is conducted by Guan et al., \cite{guan2018diagnose}, where the researchers incorporated domain knowledge into the architecture design of the network for thorax disease. The proposed network in \cite{guan2018diagnose} has three branches, one for viewing the whole image, one for viewing the local areas and one for combining the global and local information together. One of the major differences between their approach and ours is that they performed different orders to train the network while we suggested weighting the features as well as using inference rules to prioritize the features extracted.

In \cite{liu2019automated}, Liu et. al. designed a two-fold thyroid nodule classification system consisting of an ultrasound object detector to extract key features and a multi-branch convolutional neural network for classification of extracted features. In their study, the researchers utilize expert clinical knowledge to engineer feature attributes such as size and shape and place constraints on the model based on how these features are observed in practice. For example, thyroid nodule aspect ratio distributions were pre-computed based on the training set thereby ensuring detected regions would be appropriately scaled to true nodules sizes. Our proposed design is similar in that we would use multiple object detectors. However, we differ in that our object detector design uses expert knowledge to create an attention-based model due to weighting the features. Further, incorporation of inference rules creates a framework for a series of sequential object detectors, rather than simultaneous object detectors.   

% talked about the orders of detecting features regarding to inference rules
In other work conducted by Wang et al., \cite{wang2020learning}, the researchers first used a segmentation subnetwork to locate the lung area, then the lesion areas, and finally the most discriminative features \cite{xie2021survey}. In our proposed approach to leverage inference rules in image classification, the order our AI system detect features will also match the frequency they were mentioned in the think-aloud sessions (e.g. ``pleural line'' is the top mentioned feature in all think-aloud sessions and it is also listed as the first feature to look at according to the inference rules). We argue that our approach will encourage the AI model to first look at the most discriminative and supporting features, as compared to \cite{wang2020learning}. It would be interesting to further investigate how different orders of utilizing features would contribute to classification accuracy.

% we extracted more features
To the best of our knowledge, the proposed approach in \cite{guan2018diagnose}, \cite{liu2019automated}, and \cite{wang2020learning} were only tested on static Chest X-ray (CXR) or ultrasound images (not videos). Further, the features were extracted from CXR and ultrasound images. By utilizing ultrasound videos, we extracted both static and dynamic features, thus allowing for a model design capable of both object and motion detection.

% Integrating Domain Knowledge into Deep Networks for Lung Ultrasound with Applications to COVID-19
% https://ieeexplore.ieee.org/stamp/stamp.jsp?tp=&arnumber=9557273
% This paper used COVID-19 lung ultrasound data and separated features into "anatomical phenomena" (e.g., pleural line) and "sonographic artifacts" (e.g., A line, B line). Not sure if we want to discuss as well.

\section{Conclusions and Future Work}

In studies one and two, we employed Think-Aloud studies to elicit domain knowledge from physicians trained in lung ultrasound interpretation. The ultimate goal of the studies was to identify the domain knowledge that doctors use, so that we can design an AI system that incorporates that knowledge. From study one, we extracted both static and dynamic features of the ultrasound videos, from which we suggested a system design focused around object detection as well as the notion of objects across multiple video frames. In study two, we examined the reasoning process the doctors utilized to explain previously generated diagnosis. By providing correct and incorrect video labels to the doctors, we sought a method to examine how their reasoning processes changed, if at all, in the presence of incorrect prior diagnoses.

We also analyzed if different knowledge was used to generate a diagnosis (study one) vs. explain a prior diagnoses (study two). Our results show that both doctors did not mention any new features in study two compared to study one. But one doctor used inference rules when explaining the diagnoses of previously labeled ultrasound videos. In this way, we were able to prioritize the features we extracted, providing more guidance and domain knowledge for the AI system we want to design. 

% Put in future work
Our Think-Aloud studies have laid the foundation for building an AI model that can automatically diagnose Pneumothorax from ultrasound videos. We envision a model that leverages the domain knowledge we have identified to reduce the amount of training data we need and that can explain the diagnoses it generates in terms that doctors and battlefield medics can understand. Moving forward, we plan to conduct more think-aloud studies so we can generalize our model to support diagnosis of multiple medical conditions from ultrasound videos. 

%We also aim to extend this work by addressing a broad range of battlefield injuries and provide a foundation for future work.

\begin{acknowledgments}
This work was funded under the DARPA POCUS program (award \#HR00112190076). The views, opinions
and/or findings expressed are those of the author and should not be interpreted as representing the
official views or policies of the Department of Defense or the U.S. Government.
% https://www.thepocusatlas.com/, mention the name of the person who shared this
We thank Dr. Matthew Riscinti from Kinds County Emergency Medicine; ChunYi Tsai, Robert Jones Do from MetroHealth Medical Center; Francisco Norman, Hannah Kopinski (MS4) and Dr. Lindsay Davis from NYU Emergency Medicine; and Matthew Riscinti from Kings County Emergency Medicine for sharing the six open-source POCUS ATLAS videos we used in our think-aloud studies.
\end{acknowledgments}

\bibliography{references}

% \appendix

% \section{Cleaned Scripts}
% - cleaned think aloud scripts

% \section{Glossary}
% - definition of terms, put table and image

\end{document}